%% file: main.tex
\documentclass[runningheads]{llncs}

\usepackage{eccv}

\usepackage{eccvabbrv}
\let\vec\relax
\usepackage{fdsymbol}

\usepackage{graphicx}
\usepackage{booktabs}
\usepackage{pifont}
\usepackage{amsmath}
\usepackage{bbm}
\usepackage{subcaption}

\usepackage{xcolor}

\definecolor{truecyan}{RGB}{0,255,255}
\usepackage[accsupp]{axessibility}  %

\newcommand{\base}{\textsc{base}}
\newcommand{\minusaug}{\textsc{base}-no-augm}
\newcommand{\minusaffine}{\textsc{base}-no-affine}
\newcommand{\minusqueries}{\textsc{base}-same-queries}
\newcommand{\baseuniform}{\textsc{base}-uniform}
\newcommand{\siamese}{\textsc{siamese}}
\newcommand{\noselfsup}{\textsc{full}-kubric-only}
\newcommand{\nofiltering}{\textsc{base}-no-filtering}
\newcommand{\basecycle}{\textsc{base}+cycle}

\usepackage{hyperref}

\usepackage{orcidlink}

\begin{document}

\title{BootsTAP: Bootstrapped Training for Tracking-Any-Point} 

\titlerunning{BootsTAP}

\author{Carl Doersch\inst{1} \and %
Pauline Luc\inst{1} \and
Yi Yang\inst{1}\and
Dilara Gokay\inst{1} \and
Skanda Koppula\inst{1} \and
Ankush Gupta\inst{1} \and
Joseph Heyward\inst{1} \and
Ignacio Rocco\inst{1} \and
Ross Goroshin\inst{1} \and
João Carreira\inst{1} \and
Andrew Zisserman\inst{1,2}}

\authorrunning{Doersch et al.}

\institute{Google DeepMind \and
VGG, Department of Engineering Science, University of Oxford}

\maketitle

\begin{abstract}
  To endow models with greater understanding of physics and motion, it is useful to enable them to perceive how solid surfaces move and deform in real scenes. This can be formalized as Tracking-Any-Point (TAP), which requires the algorithm to track any point on solid surfaces in a video, potentially densely in space and time. Large-scale ground-truth training data for TAP is only available in simulation, which currently has a limited variety of objects and motion.  In this work, we demonstrate how large-scale, unlabeled, uncurated real-world data can improve a TAP model with minimal architectural changes, using a self-supervised student-teacher setup. We demonstrate state-of-the-art performance on the TAP-Vid benchmark surpassing previous results by a wide margin: for example, TAP-Vid-DAVIS performance improves from 61.3\% to 67.4\%, and TAP-Vid-Kinetics from 57.2\% to 62.5\%.  For visualizations, see our project webpage at \url{https://bootstap.github.io/}

\end{abstract}

\section{Introduction}
\label{sec:intro}

Despite impressive achievements in the vision and language capability of generalist AI systems, physical and spatial reasoning remain notable weaknesses of state-of-the art vision models~\cite{gpt4v, gemini}. This limits their application in many domains like robotics, video generation, and 3D asset creation -- all of which require an understanding of the complex motions and physical interactions in a scene.
Tracking-Any-Point (TAP)~\cite{doersch2022tap} is a promising approach to represent precise motions in videos, and recent work has demonstrated compelling usage of TAP in robotics~\cite{vecerik2023robotap,abbeel,bharadhwaj2024track2act}, 3D reconstruction~\cite{wang2023visual}, video generation~\cite{doersch2023tapir}, and video editing~\cite{yu2023videodoodles}. In TAP, algorithms are fed a video and a set of query points---potentially densely across the video---and must output the tracked location of these query points in the video's other frames. If the point is not visible in a frame, the point is marked as occluded in that frame. This approach has many advantages: it is a highly general task, as correspondences for surface points are well-defined for opaque, solid surfaces, and it provides rich information about the deformation and motion of objects across long time periods.

\begin{figure}[t]
  \centering
  \includegraphics[width=0.98\linewidth]{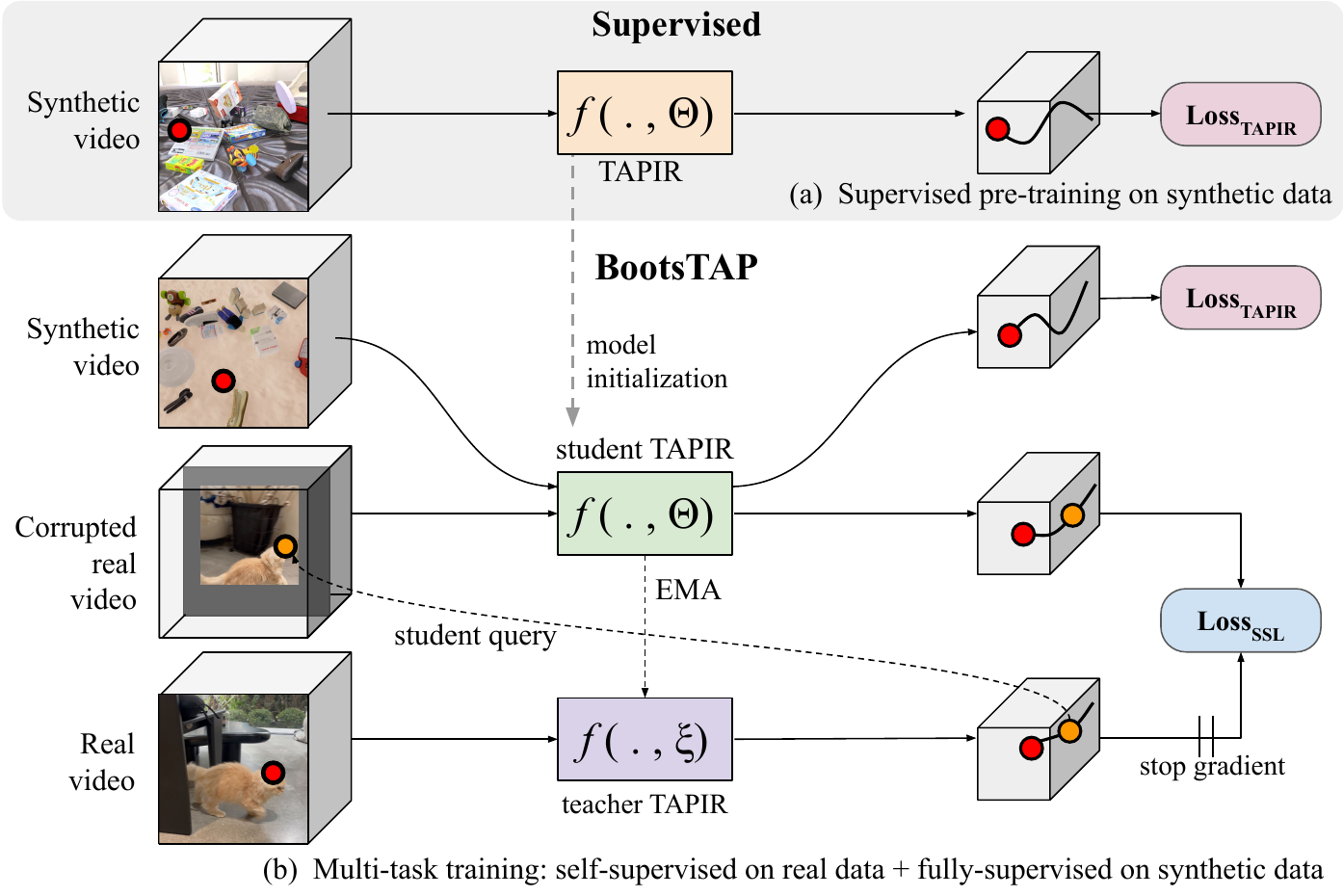}
\vspace{-2mm}
  \caption{\textbf{Bootstrapped training for tracking-any-point.}  After initializing a TAPIR model with standard supervised training, we bootstrap the model on real data by adding an additional self-supervised loss.  We apply a teacher model (a simple EMA of the student model) to get pseudo-ground-truth labels for a video.  We then apply spatial transformations and corruptions to the video to make the task harder for the student, and train the student to reproduce the teacher's predictions from any query point along the teacher's trajectory.}
  \label{fig:teaser}
  \vspace{-2mm}
\end{figure}

The main challenge for building TAP models, however, is the lack of training data: in the real world, we must rely on manual labeling, which is arduous and imprecise~\cite{doersch2022tap}, or on 3D sensing~\cite{balasingam2023drivetrack}, which is only available in limited scenarios and quantity. Thus, state-of-the-art methods have relied on synthetic data~\cite{greff2022kubric,zheng2023pointodyssey}. In this work, however, we overcome this limitation and demonstrate that unlabeled real-world videos can be used to improve point tracking, using self-consistency as a supervisory signal.  In particular, we know that when tracks are correct for a given video, then 1) spatial transformations of the video should result in an equivalent spatial transformation of the trajectories, 2) that different query points along the same trajectory should produce the same track, and 3) that non-spatial data augmentation (e.g. image compression) should not affect results.  Deviations from this can be treated as an error signal for learning.

Our architecture is outlined in Figure~\ref{fig:teaser}.  We begin with a strong ``teacher'' model pre-trained using supervised learning on synthetic data (in our case, a TAPIR~\cite{doersch2023tapir} model) which serves as initialization for both a ``teacher'' and a ``student'' model.  Given an unlabeled input video, we make a prediction using the teacher model, which serves as pseudo-ground-truth for the student.  We then generate a second ``view'' of the video by applying affine transformations that vary smoothly in time, re-sampling frames to a lower resolution, and adding JPEG corruption, and padding back to the original size.  We input the second view to the ``student'' network and use a query point sampled from the teacher's prediction (transformed consistently with the transformation applied to the video). The student's prediction is then transformed back into the original coordinate space. We then use a self-supervised loss (SSL) to update the student's weights: that is, we apply TAPIR's original loss function to the student predictions, using the teacher's predictions as pseudo-ground-truth. The teacher's weights are updated by using an exponential moving average (EMA) of the student's weights.  We take steps to ensure that the teacher's predictions used for training are more likely to be accurate than the student's: (i) the corruptions that degrade and downsample the video are only applied to the student's inputs, (ii) we use an EMA of the student's weights as the teacher's weights, a common trick for stabilizing student-teacher learning~\cite{grill2020bootstrap,tarvainen2017mean}. %
Co-training using this formulation on real-world videos, in addition to training on synthetic data, provides a substantial boost over prior state-of-the-art across the entire TAP-Vid benchmark.

\vspace{1pt}
\noindent In summary, our contributions are as follows:
\vspace{-5pt}
\begin{enumerate}
    \item We demonstrate the first large-scale pipeline for improving video point tracking using a large dataset of unannotated videos, based on straightforward properties of real trajectories: (i) predictions should vary consistently with spatial transformations of the video, and (ii) predictions should be invariant to the choice of query point along a given trajectory.
    \item We analyze the importance of varying model components, and show that a surprisingly simple formulation is sufficient to achieve good results.
    \item We show that the resulting formulation achieves new SOTA results on point tracking benchmarks, while requiring minimal architectural changes.
    \item We will release a model and checkpoint on GitHub, including model implementations in both JAX and PyTorch for the community to use.
\end{enumerate}

\section{Related Work}
\paragraph{Tracking-Any-Point.} The ability to track densely-sampled points over long video sequences is a generic visual capability \cite{rubinstein2012towards, sand2008particle}. Because this visual task provides a rich output that is well-defined independent of semantic or linguisitic categories (unlike classification, detection, and semantic segmentation), it is more generically useful and can support other visual capabilities like video editing~\cite{yu2023videodoodles}, 3D estimation~\cite{wang2023omnimotion}, object segmentation \cite{ochs2013segmentation,rajivc2023segment}, camera tracking~\cite{chen2024leap} and even robotics \cite{vecerik2023robotap,abbeel}. Point tracking has recently experienced a flurry of recent works including new datasets \cite{doersch2022tap, zheng2023pointodyssey, balasingam2023drivetrack} and algorithms \cite{harley2022pips, karaev2023cotracker, doersch2023tapir, wang2023omnimotion, neoral2024mft, bian2024context, lemoing2023dense}.
Current state-of-the-art works mainly train in a supervised manner, relying heavily on synthetic data~\cite{greff2022kubric,zheng2023pointodyssey} which has a large domain gap with the real world.

\paragraph{Self-supervised correspondence via photometric loss.} Tracking has long been a target of self-supervised learning due to the lack of reliable supervised data, especially at the point level. A wide variety of proxy supervisory signals have been proposed, all with their own limitations. Photometric losses use reconstruction, and are particularly popular in optical flow, but occlusions, lighting changes, and repeated (or constant) textures, typically result in multiple or false appearance matches. To compensate for this, these methods typically rely on complicated priors such as multi-frame estimation \cite{janai2018unsupervised}, explicit occlusion handling \cite{stone2021smurf, wang2018occlusion}, improved data augmentation \cite{liu2020learning}, additional loss terms  \cite{liu2019ddflow,liu2019selflow, meister2018unflow}, and robust loss functions which avoid degenerate solutions \cite{yu2016back, ren2017unsupervised, marsal2023brightflow}. Methods that combine feature learning with appearance reconstruction, such as~\cite{vondrick2018tracking, lai2019self, Lai20}, have demonstrated long-range tracking. Matches based on local appearance are more likely to correspond to motion in high resolution videos because they are able to resolve detailed textures \cite{janai2017slow}; we make use of this observation in our work.

\paragraph{Temporal continuity and cycle-consistency.} Other works use images or videos to perform more general feature learning, with the aim that features in correspondence should be more similar than those which are not. Temporal continuity in videos has long been used to obtain such correspondences \cite{foldiak1991learning, wiskott2002slow, goroshin2015unsupervised, wang2015unsupervised, jabri2020space}, resulting in features which have proven to be effective for object tracking \cite{denil2012learning, goroshin2015learning}. Temporal cycle-consistency~\cite{wang2019learning,Bian_2022_CVPR} can also result in features useful for tracking; however this learning method fails to provide useful supervision in challenging situations such as occlusions. 

\paragraph{Semi-supervised correspondence.} A final self-supervised approach is to create pseudo-ground-truth correspondences for semi-supervised training~\cite{huang2023self, shen2023self}. Such approaches have a long history in optical flow \cite{liu2019ddflow, liu2019selflow,im2022semi,novak2020new}, although with mixed results, typically requiring complex training setups such as GANs~\cite{lai2017semi} or connecting the student to the teacher~\cite{liu2021learning} to prevent trivial solutions. They have only been applied to longer-term point tracking more recently~\cite{wang2023omnimotion,sun2024refining}. OmniMotion computes initial point tracks using RAFT~\cite{teed2020raft} or TAP-Net~\cite{doersch2022tap} and infers a full pseudo-3D interpretation of the scene in the form of a neural network. Although this method improves point tracks compared to their initialization, it never retrains a general TAP model on the self-labeled data.  Perhaps most related is Li et al.~\cite{li2023learning}, which proposes a self-supervised loss based on reconstruction, in addition to supervised point tracking loss and an adversarial domain adaptation loss.  The final algorithm is complex, and performs far below our work (59.8 on TAP-Vid-DAVIS $< \delta^{x}_{avg}$, versus 78.1 for our work), with the self-supervised providing a relatively small boost.  Concurrent work~\cite{sun2024refining}, on the other hand, saves a dataset of point tracks and retrains the underlying model on them, using data augmentations similar to ours. We discuss the differences in detail in the following section, after presenting our approach.

\section{Method}

When developing a self-supervised training method for TAP, it is important to note that TAP has a precise, correct answer for almost every query point.  This is different from typical visual self-supervised learning, where the representation can be arbitrary, as long as semantically similar images have similar representations.  Supervised learning on synthetic data provides a strong initial guess in many situations, but care must be taken to ensure that the self-supervised algorithm does not find ``trivial shortcuts''~\cite{doersch2015unsupervised} that become self-reinforcing and harm the initialization.  %

\begin{figure}[t]
  \centering
  \includegraphics[width=\linewidth]{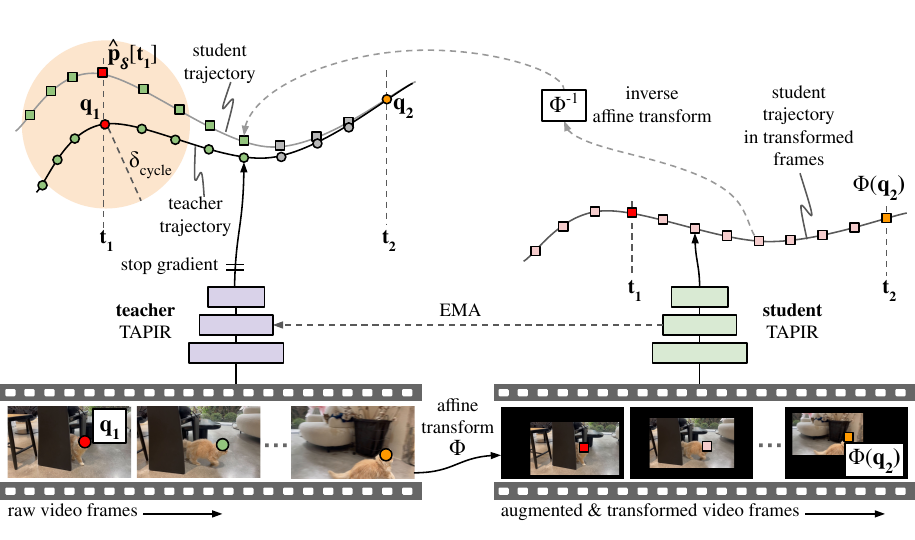}
  \caption{\textbf{Bootstrapped training for Tracking-Any-Point.}
The teacher TAPIR produces a pseudo-label trajectory from query point $q_1$ at time $t_1$. Video frames undergo affine transformations $\Phi$ that vary smoothly in time and are augmented with JPEG artifacts, then fed to the student TAPIR, which predicts a trajectory from query point $\Phi(q_2)$ at time  $t_2$ (sampled from the teacher's prediction, then transformed to the student video space using $\Phi$). The student trajectory is transformed back, and loss is computed against the teacher's trajectory.
To maximize the chances that we train on accurate trajectories, we remove trajectories where the student's prediction at time $t_1$ is too far from the teacher query point $q_1$ (i.e. not cycle-consistent; light-orange disk).
}
  \label{fig:overview}
\end{figure}

Our formulation relies on two facts about point tracks that are true for points on any solid, opaque surface. 
First, spatial transformations (e.g.\ affine transformations) which are applied to the video will result in equivalent spatial transformations of the point tracks (i.e.\ the tracks are ``equivariant'' under spatial transformation), while the tracks are invariant to many other factors of variation that do not move the image content (e.g.\ color changes, noise).
Second, the algorithm should output the same track regardless of which point along the track is used as a query; mathematically, this means that each trajectory forms an equivalence class.  %
One could imagine enforcing the desired equivariance and invariance properties using a simple Siamese-network formulation~\cite{hadsell2006dimensionality}, where a single network is trained to output consistent predictions on two different `views' of the data (i.e., augmented and transformed versions of the video and tracks). 
However, we find that minimizing the difference between the two outputs---and backpropping both---results in predictions degrading toward trivial solutions (e.g. over-smoothing of tracks, or tracking the image boundary instead of the image content).  In fact, the model can learn to distinguish between synthetic and real data resulting in trivial solutions on the real, unlabeled data only. To prevent this, we adopt a student-teacher framework, where the student's view of the data is made more challenging by augmentations, and the teacher does not receive gradients that may corrupt its predictions.
Figure~\ref{fig:overview} shows the overall pipeline.  

\paragraph{Loss functions.} We start with a baseline TAPIR network pre-trained on Kubric following~\cite{doersch2023tapir}.  
Let $\hat{y}=\{\hat{p},\hat{o},\hat{u}\}$ be the predictions: $\hat{p}\in \mathbb{R}^{T \times 2}$ is the position, $\hat{o}\in \mathbb{R}^{T}$ is an occlusion logit, and $\hat{u}\in \mathbb{R}^{T}$ is an uncertainty logit, where $T$ is the number of frames. Calling $p[t]$ and $o[t]$ the ground truth targets for frame $t$, recall that the standard TAPIR loss for a single trajectory is defined as:

\begin{align}
        \mathcal{L}_{tapir}(\hat{p}[t], \hat{o}[t], \hat{u}[t]) = & \hspace{.3em} \text{Huber}(\hat{p}[t], p[t]) (1 - o[t]) & \mbox{Position loss} \nonumber\\
             & + \text{BCE}(\hat{o}[t], o[t]) & \mbox{Occlusion loss} \\
             & + \text{BCE}(\hat{u}[t], u[t])(1 - o[t]) & \mbox{Uncertainty loss} \nonumber
\end{align}

\noindent where Huber is the Huber loss and BCE is the sigmoid binary cross-entropy.  The target for the uncertainty logit is defined as $u[t] = \mathbbm{1}(d(p[t], \hat{p}[t]) > \delta)$, where $d$ the $L_2$ distance and $\delta$ is a threshold on the distance, set to 6 pixels, and $\mathbbm{1}$ is an indicator function.  That is, the uncertainty loss trains the model to predict whether its own prediction is likely to be within a threshold of the ground truth.  

After pre-training, we add extra capacity to the model to absorb the extra training data: 5 layers of 2D conv-residual layers to the backbone with a channel multiplier of 4, which roughly doubles the number of parameters in the backbone (see Appendix~\ref{app:convnet} for details).  These are initialized to the identity following ``zero init''~\cite{goyal2017accurate}.  
Let $\hat{y}_{\mathcal{S}}=\{\hat{p}_{\mathcal{S}},\hat{o}_{\mathcal{S}},\hat{u}_{\mathcal{S}}\}$ now refer to the student predictions. We derive pseudo-labels $y_{\mathcal{T}}=\{p_{\mathcal{T}},o_{\mathcal{T}},u_{\mathcal{T}}\}$ from the teacher's predictions $\hat{y}_{\mathcal{T}}=\{\hat{p}_{\mathcal{T}},\hat{o}_{\mathcal{T}},\hat{u}_{\mathcal{T}}\}$ as follows:

\begin{equation}
    p_{\mathcal{T}}[t] = \hat{p}_{\mathcal{T}}[t] \quad ; \quad 
    o_{\mathcal{T}}[t] = \mathbbm{1}(\hat{o}_{\mathcal{T}}[t] > 0); \quad 
    u_{\mathcal{T}}[t] = \mathbbm{1}(d(\hat{p}_{\mathcal{T}}[t], \hat{p}_{\mathcal{S}}[t]) > \delta)
\end{equation}

\noindent where $t$ indexes time.  The loss $\ell_{ssl}(\hat{p}_{\mathcal{S}}[t], \hat{o}_{\mathcal{S}}[t], \hat{u}_{\mathcal{S}}[t])$ for a given video frame $t$ is derived from the TAPIR loss, treating the pseudo-labels as ground-truth, and defined as:

\begin{align}
        \ell_{ssl}(\hat{p}_{\mathcal{S}}[t], \hat{o}_{\mathcal{S}}[t], \hat{u}_{\mathcal{S}}[t]) = & \hspace{.3em} \text{Huber}(\hat{p}_{\mathcal{S}}[t], p_{\mathcal{T}}[t]) (1 - o_{\mathcal{T}}[t]) \nonumber\\
            & + \text{BCE}(\hat{o}_{\mathcal{S}}[t], o_{\mathcal{T}}[t])  \\
            & + \text{BCE}(\hat{u}_{\mathcal{S}}[t], u_{\mathcal{T}}[t])(1 - o_{\mathcal{T}}[t]) \nonumber
\end{align}

\noindent Note that TAPIR's loss uses multiple refinement iterations, but we always use the teacher's final prediction to derive pseudo-ground-truth; therefore, refined predictions serve as supervision for unrefined ones, encouraging stronger features that enable faster convergence.  

\paragraph{Video degradations.} While the above formulation is well-defined, if the student and teacher both receive the same video and query point, we expect the loss to be trivially close to zero; therefore, we apply transformations and corruptions to the student's view of the video. 
Given an input video, we create a second view by resizing each frame to a smaller resolution $r$ and superimposing it onto a black background at a random position $(h,w)$ within this background. $r$ varies linearly over time, meaning that the frames gradually become larger or smaller within the fixed-size black background. Overall, the decreased resolution degrades the student view, and this increases task difficulty for the student. The location of these frames also move with time, and $(h,w)$ follows a linear trajectory within the black background. Formally, this is a frame-wise axis-aligned affine transformation $\Phi$ on coordinates, applied to the pixels. 
We also apply $\Phi$ to the student query coordinates. 
We further degrade this view by applying a random JPEG degradation to make the task more difficult, before pasting it onto the black background. Both operations lose texture information; therefore, the network must learn higher-level---and possibly semantic---cues (e.g.\ the tip of the top left ear of the cat), rather than lower-level texture matching in order to track points correctly. 
We apply the inverse affine transformation $\Phi^{-1}$ to map the student's predictions back to the original input coordinate space, before feeding these to the loss.
We describe these transformations and corruptions in more detail in  Appendix~\ref{appendix:affine}. 

\paragraph{Choosing the sample point.} We enforce that each trajectory forms an equivalence class 
by training the model to produce the same track regardless of which point is used as a query.
While we do not have access to the ground-truth trajectories to sample different query points from, we can use the teacher model's predictions to form pairs of query points.
First, we sample a query point $Q_1=(q_1,t_1)$, where $q_1$ is an $(x, y)$ coordinate, and $t_1$ is a frame index, both sampled uniformly.
Then the student's query is sampled randomly from the teacher's trajectory, i.e. $ Q_2 = (q_2,t_2) \in \{ (p_{\mathcal{T}}[t],t) ; t \textit{ s.t. } o_{\mathcal{T}}[t] = 0\}$.

Note, however, that if the teacher has not tracked the point correctly, the student's query might be a different real-world point than the teacher's, leading to an erroneous training signal.  
To prevent this, we use cycle-consistency of the student and teacher trajectories, and ignore the loss for trajectories that don't form a valid cycle, as depicted by the orange circle in Figure~\ref{fig:overview}.  Formally, we implement this as a mask defined as:

\begin{equation}
    m_{cycle} =  \mathbbm{1} \left( d(\hat{p}_{\mathcal{S}}[t_1], q_1) < \delta_{cycle} \right) \quad * \quad \mathbbm{1} \left(\hat{o}_{\mathcal{S}}[t_1] \le 0\right)
\end{equation}

Here, $\delta_{cycle}$ is a distance threshold hyperparameter, which we set to 4 pixels.

Note that there is a special case when the student and teacher have the same query point: there is no longer any uncertainty regarding whether the point is on the same trajectory.  These points are reliable while also being less challenging.  
We compromise between extremes, and sample $Q_1=Q_2$ with probability 0.5, and sample with equal probabilities from the remaining visible points in the teacher prediction.
The final self-supervised loss for a single trajectory is then:

\begin{equation}
        \mathcal{L}_{SSL}
            = \sum_t
            m_{cycle}^{t} * \ell_{ssl}^{t} %
\end{equation}

In practice, we sample 128 query points per input video and average the loss for all of them.
We provide pseudocode for the algorithm in Appendix~\ref{app:recap}.

To avoid catastrophic forgetting, we continue training on the Kubric dataset with the regular supervised TAPIR loss.
Our training setup follows prior work on multi-task self-supervised learning~\cite{doersch2017multi}: we maintain separate Adam optimizer parameters to compute separate updates for both tasks, and then apply the gradients with their own learning rates.  As the self-supervised task is more expensive due to the extra forward pass, we use half the batch size for self-supervised updates, and therefore we halve the learning rate for these updates. See Appendix~\ref{appendix:details} for more details.

\vspace{2mm}
\noindent \textbf{Differences between our approach and~\cite{sun2024refining}.} Concurrent work reproduces some of these decisions, including using cycle-consistency as a method of filtering and using affine transformations when augmenting the student view.  However, there are a few key differences. First, rather than a student-teacher setup, they compute trajectories only once and freeze the training data, meaning that the model is permanently trained to reproduce errors in the original labeling. Furthermore, the work fine-tunes on the target dataset, meaning that transfer to a new domain may require a large training set in that domain on which to fine-tune; in contrast, our work demonstrates that it's possible to train on a single large dataset that covers many domains, meaning that fine-tuning is unnecessary. 

\section{Experiments}
We train our model on over 15 million 24-frame clips from publicly-available online videos, in conjunction with standard training on Kubric.  The resulting model is essentially a drop-in replacement for TAPIR (albeit with slightly larger computational requirements due to the extra layers).  We evaluate on the TAP-Vid benchmark using the standard protocol.

\subsection{Training datasets}

We collected a video dataset from publicly accessible videos selected from categories that typically contain high-quality and realistic motion (such as lifestyle and one-shot videos). Conversely, we omitted videos from categories with low visual complexity or unrealistic motions, such as tutorial videos, lyrics videos, and animations. To maintain consistency, we exclusively obtained videos shot at 60fps. Additionally, we applied a quality metric by only considering videos with over 200 views.
We removed the first and last 2 seconds of each video, as these often contain intros and outros with text or other overlays. From each video, we randomly sampled five clips, excluding those with overlay/watermarked frames, which were identified by checking the horizontal and vertical gradients and computing the pixel-wise median (similar to~\cite{dekel2017effectiveness}). Furthermore, we expect the teacher signal will be more reliable on continuous shots due to temporal continuity; therefore, clips with shot boundary changes are detected and removed based on~\cite{boreczky1996comparison, yusoff2000video, truong2000new, mas2003video} with additional accuracy improvements based on full-frame geometric alignment. In total, we generated 15 million clips for training. 

\subsection{Evaluation datasets}

We rely on the TAP-Vid~\cite{doersch2022tap} and RoboTAP~\cite{vecerik2023robotap} benchmarks for quantitative evaluation; in all cases, we evaluate zero-shot on the entire benchmark, resizing to $256 \times 256$ before evaluating according to the standard procedure~\cite{doersch2022tap}. This consists of five datasets: \textbf{TAP-Vid-Kinetics} which contains online videos of human actions and may include cuts~\cite{carreira2017quo}; \textbf{TAP-Vid-DAVIS} which is based on the DAVIS object tracking benchmark~\cite{perazzi2016benchmark}; \textbf{TAP-Vid-RGB-Stacking} which contains synthetic tracks for videos of robotic manipulation which have little texture; and \textbf{RoboTAP} which contains real-world robotic manipulation videos~\cite{vecerik2023robotap}, all of which include ground truth.  Evaluation is performed by measuring occlusion accuracy (OA), $\mathbf{< \delta^{x}_{avg}}$ which measures the fraction of point estimates within a specified distance to the ground truth location, averaged across 5 thresholds, and Average Jaccard (AJ) which measures a combination of these two.  There are two dataset querying ``modes'': \emph{query first} (\emph{q\_first}) uses the first visible point on each trajectory as a query, while \emph{strided} uses every fifth point along the trajectory as a separate query. We also include qualitative evaluations on two robotics datasets without ground truth: \textbf{RoboCAT-NIST}, a subset of the data collected for RoboCat~\cite{bousmalis2023robocat}, and \textbf{Libero}~\cite{liu2024libero}, a dataset where point tracking has already proven useful for robotic manipulation~\cite{abbeel}. See Appendix~\ref{app:evals} for details on these datasets and metrics.  

\subsection{Results}

\begin{table}[tb]
  \setlength{\tabcolsep}{3pt}
  \caption{Comparison of performance on the TAP-Vid datasets.  AJ (Average Jaccard; higher is better) measures both occlusion and position accuracy.  $<\delta^{x}_{avg}$ (higher is better) measures only localization performance, ignoring occlusion accuracy.  OA (Occlusion Accuracy; higher is better) measures only accuracy in predicting occlusion.
  }
  \label{tab:tapir_perf}
  \centering
  \begin{tabular}{lccc|ccc|ccc}
    \toprule
 & \multicolumn{3}{c|}{Kinetics} & \multicolumn{3}{c|}{DAVIS} & \multicolumn{3}{c}{RGB-Stacking} \\
Method &  AJ & $<\delta^{x}_{avg}$ & OA &  AJ & $<\delta^{x}_{avg}$ & OA &  AJ & $<\delta^{x}_{avg}$ & OA \\
\midrule
COTR~\cite{jiang2021cotr} & 19.0  & 38.8  & 57.4  &   35.4 & 51.3 & 80.2 & 6.8 & 13.5 & 79.1 \\ %
Kubric-VFS-Like~\cite{greff2022kubric} & 40.5 & 59.0 & 80.0 & 33.1 & 48.5 & 79.4 & 57.9 & 72.6 & 91.9 \\ %
RAFT~\cite{teed2020raft} & 34.5 & 52.5 & 79.7 &  30.0 & 46.3 & 79.6 & 44.0 & 58.6 & 90.4 \\ %
TAP-Net~\cite{doersch2022tap} & 46.6 & 60.9 & 85.0 &  38.4 & 53.1 & 82.3 & 59.9 & 72.8 & 90.4 \\ %
PIPs~\cite{harley2022pips} & 35.3 & 54.8 & 77.4 & 42.0 & 59.4 & 82.1 & 37.3 & 51.0 & 91.6 \\ %
PIPs+Refinement~\cite{sun2024refining} & - & - & - & 42.5 & 60.0 & - & - & - & - \\ 
TAPIR~\cite{doersch2023tapir} & 57.2 & 70.1 & 87.8 &  61.3 & 73.6 & 88.8 & 62.7 & 74.6 & 91.6 \\ %
CoTracker~\cite{karaev2023cotracker} & 57.3 & 70.6 & 87.5 &  64.8 & \textbf{79.1} & 88.7 & 65.9 & 80.6 & 85.0  \\ %

\midrule
BootsTAPIR & \textbf{61.4} & \textbf{74.2} & \textbf{89.7} & \textbf{66.2} & 78.1 & \textbf{91.0} & \textbf{72.4} & \textbf{83.1} & \textbf{91.2} \\
  \bottomrule
  \end{tabular}
\end{table}

Our results are shown in Table~\ref{tab:tapir_perf}.  Note that all of our numbers come from a single checkpoint, which has not seen the relevant datasets.  Relative to our base architecture, our bootstrapping approach provides a substantial gain across all metrics.  We also outperform CoTracker on DAVIS, though this is due more to improvements in occlusion accuracy than position accuracy.  This is despite TAPIR having a simpler architecture than CoTracker, which requires cross attention to other points which must be chosen with a hand-tuned distribution, whereas TAPIR tracks points independently.  CoTracker results are also obtained by upsampling videos to $384\times 512$, which further increases compute time, whereas ours are computed directly on $256 \times 256$ videos.

\begin{table}[tb]
  \setlength{\tabcolsep}{5pt}
  \caption{Comparison of performance under query-first metrics for Kinetics, TAP-Vid DAVIS, and RoboTAP (standard for this dataset).
  }
  \label{tab:tapir_perf_first}
  \centering
  \begin{tabular}{lccc|ccc|ccc}
    \toprule
 & \multicolumn{3}{c|}{Kinetics} & \multicolumn{3}{c|}{DAVIS} & \multicolumn{3}{c}{RoboTAP} \\ %
Method & AJ & $<\delta^{x}_{avg}$ & OA & AJ & $<\delta^{x}_{avg}$ & OA & AJ & $<\delta^{x}_{avg}$ & OA \\
\midrule
TAP-Net~\cite{doersch2022tap} & 38.5 & 54.4 & 80.6 & 33.0 & 48.6 & 78.8 & 45.1 & 62.1 & 82.9 \\
TAPIR~\cite{doersch2023tapir} & 49.6 & 64.2 & 85.0 & 56.2 & 70.0 & 86.5 & 59.6 & 73.4 & 87.0 \\
CoTracker~\cite{karaev2023cotracker} & 48.7 & 64.3 & \textbf{86.5} & 60.6 & \textbf{75.4} & \textbf{89.3} & 54.0 & 65.5 & 78.8 \\
\midrule
BootsTAPIR & \textbf{54.6} & \textbf{68.4} & \textbf{86.5} & \textbf{61.4} & 73.6 & 88.7 & \textbf{64.9} & \textbf{80.1} & \textbf{86.3} \\
\bottomrule
  \end{tabular}
\end{table}

Table~\ref{tab:tapir_perf_first} shows performance under \textit{q\_first} mode.  Here, we see that bootstrapping outperforms prior works by a wide margin on Kinetics; this is likely because TAPIR's global search is more robust to large occlusions and cuts, which are more prominent in Kinetics.  This search might harm performance in datasets like DAVIS with a stronger temporal continuity bias.
Perhaps most impressive is the strong improvement in RoboTAP--over 5\% absolute performance--despite RoboTAP looking very different from typical online videos.  We see similar results for RGB-Stacking in Table~\ref{tab:tapir_perf}.  These two datasets have large textureless regions; such regions are challenging to track without object-aware priors, which are difficult to obtain from synthetic datasets.

Figure~\ref{fig:davis_robotap} shows qualitative examples of some cases where BootsTAPIR improves performance.  We see improvements on examples where texture cues are ambiguous (e.g. the dark jacket and trousers) where prior knowledge of common object shape can improve performance, as well as points near object boundaries (e.g. the dog's ears) where a model trained on synthetic data with different appearance may struggle to estimate the correct segmentation.  We also note that BootsTAPIR improves on many cases where TAPIR marks a point as occluded even when it is still visible, such as the person's arm.  On RoboTAP, the model improves on occlusion estimation for the textureless gripper. It also deals well with large changes in scale as the gripper approaches the shoe, as well as shiny objects, both of which are less common in Kubric. Our project webpage \url{https://bootstap.github.io/} includes video examples, which makes the improvements more obvious.

\newcommand{\myimagewidth}{0.15\textwidth}
\newcommand{\mydavisimagewidth}{0.267\textwidth}
\newcommand{\myarmimagewidth}{0.24\textwidth}
\newcommand{\myroboimagewidth}{0.2\textwidth}
\begin{figure}[h]
\centering
\begin{tabular}{cccccc}
\includegraphics[width=\mydavisimagewidth]{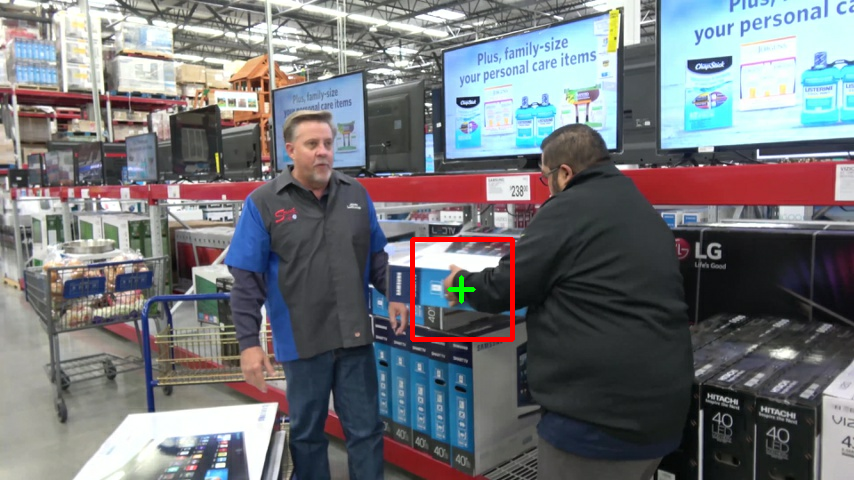} & \includegraphics[width=\myimagewidth]{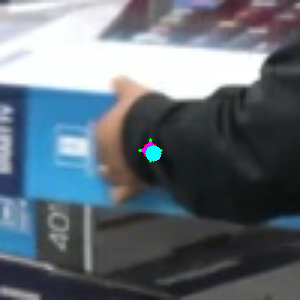} & \includegraphics[width=\myimagewidth]{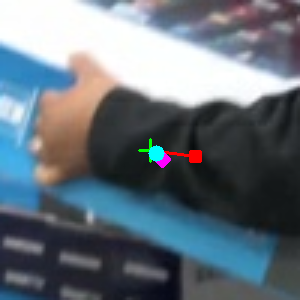} & \includegraphics[width=\myimagewidth]{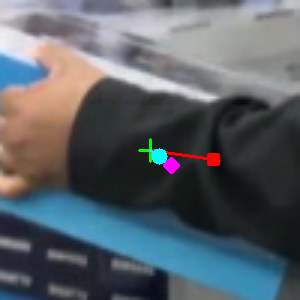} & \includegraphics[width=\myimagewidth]{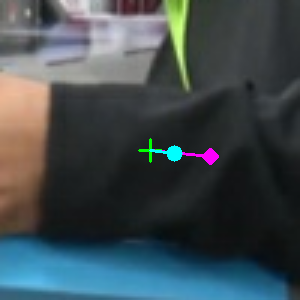} \\
\includegraphics[width=\mydavisimagewidth]{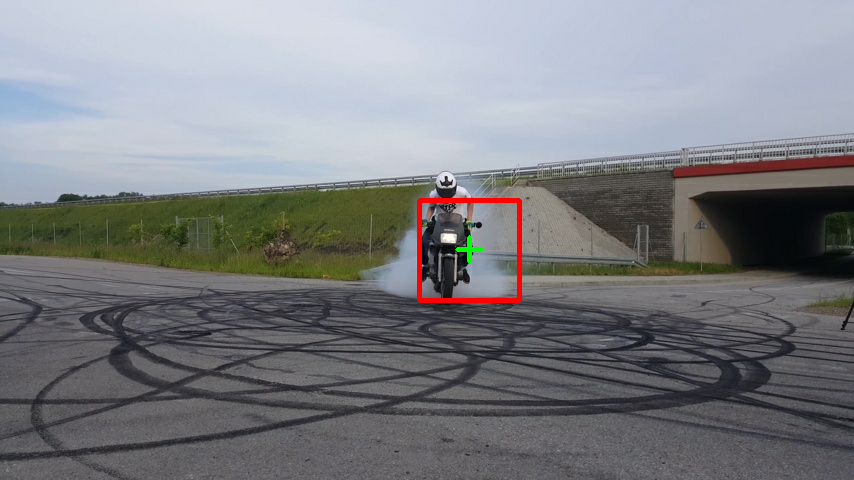} & \includegraphics[width=\myimagewidth]{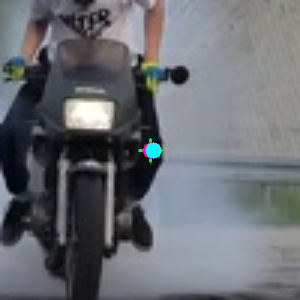} & \includegraphics[width=\myimagewidth]{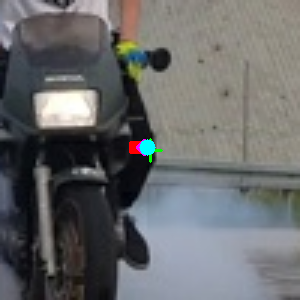} & \includegraphics[width=\myimagewidth]{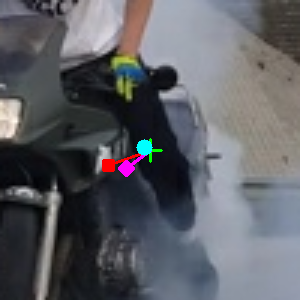} & \includegraphics[width=\myimagewidth]{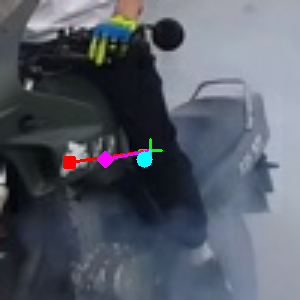} \\
\includegraphics[width=\mydavisimagewidth]{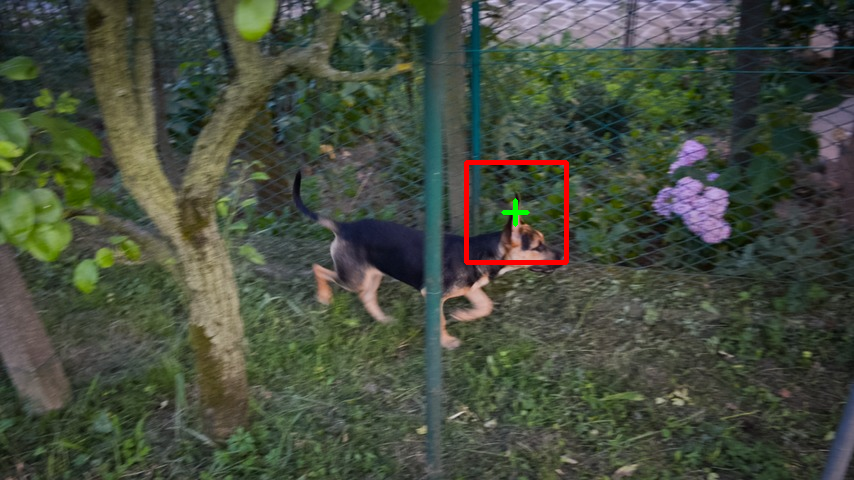} & \includegraphics[width=\myimagewidth]{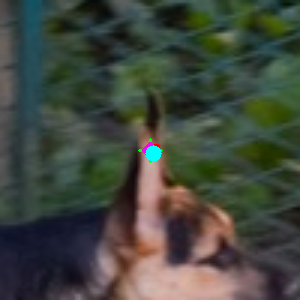} & \includegraphics[width=\myimagewidth]{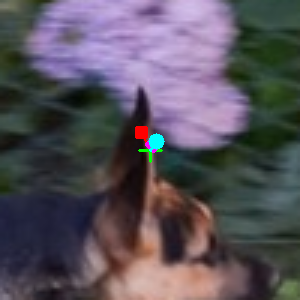} & \includegraphics[width=\myimagewidth]{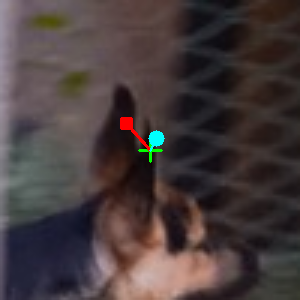} & \includegraphics[width=\myimagewidth]{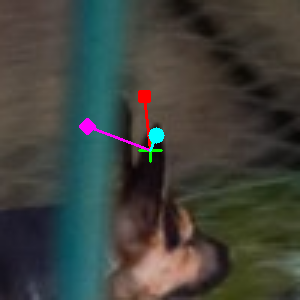} \\
\includegraphics[width=\myarmimagewidth]{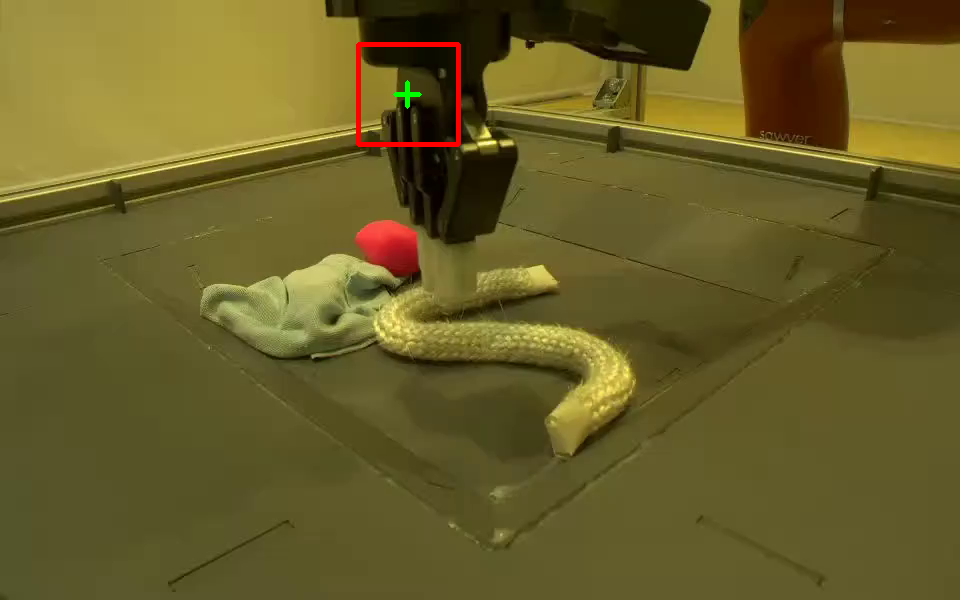} & \includegraphics[width=\myimagewidth]{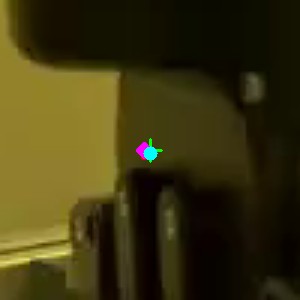} & \includegraphics[width=\myimagewidth]{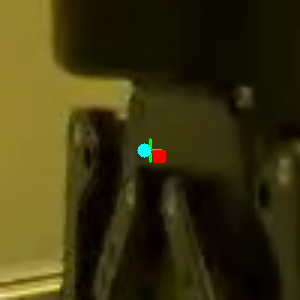} & \includegraphics[width=\myimagewidth]{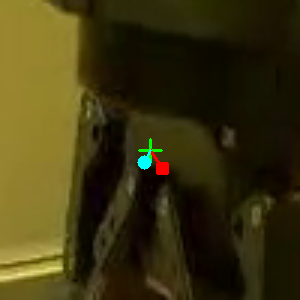} & \includegraphics[width=\myimagewidth]{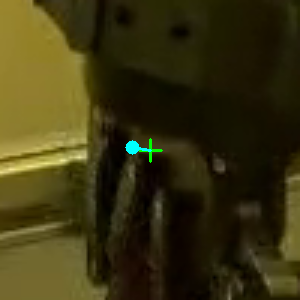} \\
\includegraphics[width=\myroboimagewidth]{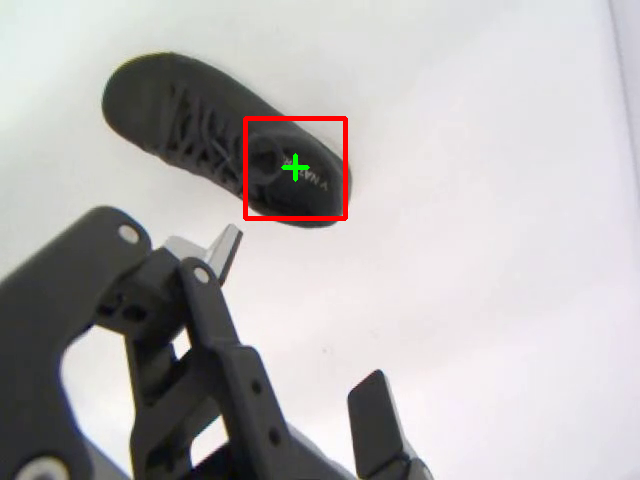} & \includegraphics[width=\myimagewidth]{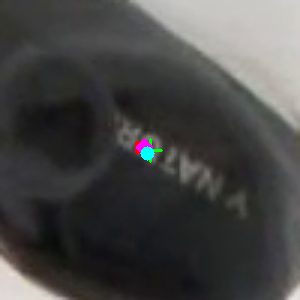} & \includegraphics[width=\myimagewidth]{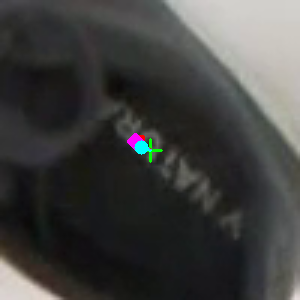} & \includegraphics[width=\myimagewidth]{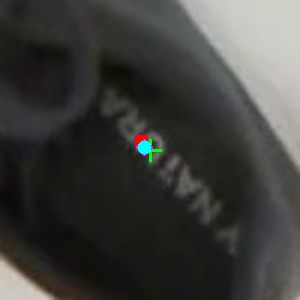} & \includegraphics[width=\myimagewidth]{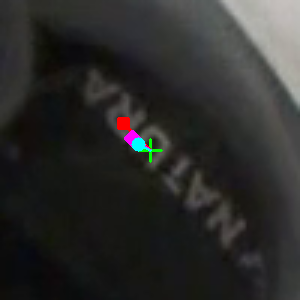} \\
\includegraphics[width=\myroboimagewidth]{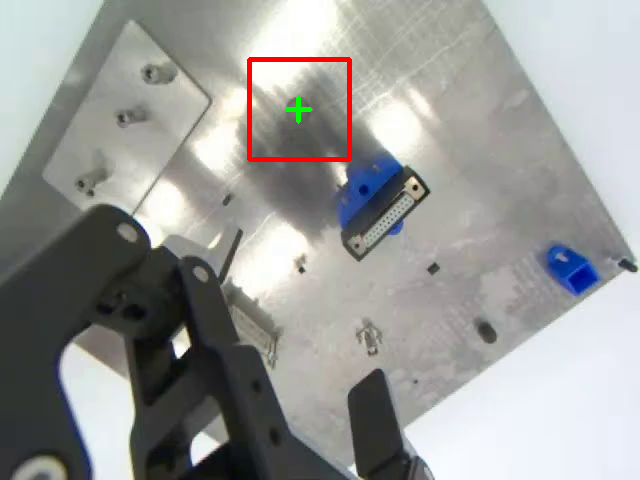} & \includegraphics[width=\myimagewidth]{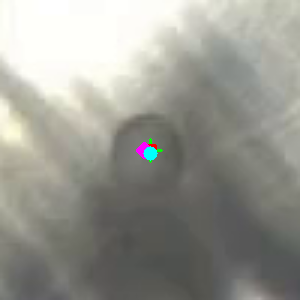} & \includegraphics[width=\myimagewidth]{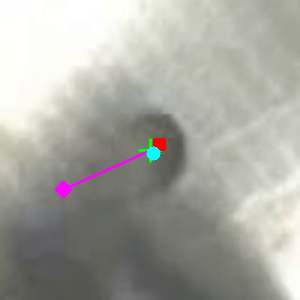} & \includegraphics[width=\myimagewidth]{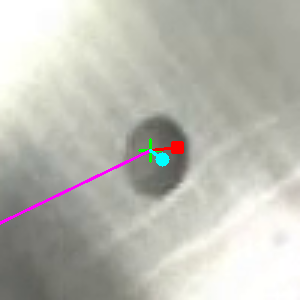} & \includegraphics[width=\myimagewidth]{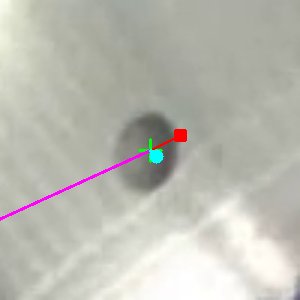} \\
\end{tabular}
\caption{Comparison between TAPIR ({\color{red}$\medblacksquare$}), Cotracker ({\color{magenta}$\medblackdiamond$}) and BootsTAPIR ({\color{truecyan}$\medblackcircle$}), and the ground-truth points ({\color{green}\textbf{+}}) on TAP-Vid-DAVIS and RoboTAP benchmarks. We show the initial query frame, and a closeup of four later frames.}
\label{fig:davis_robotap}
\end{figure}

\begin{figure}[tb]
  \centering
  \includegraphics[width=.95\linewidth]{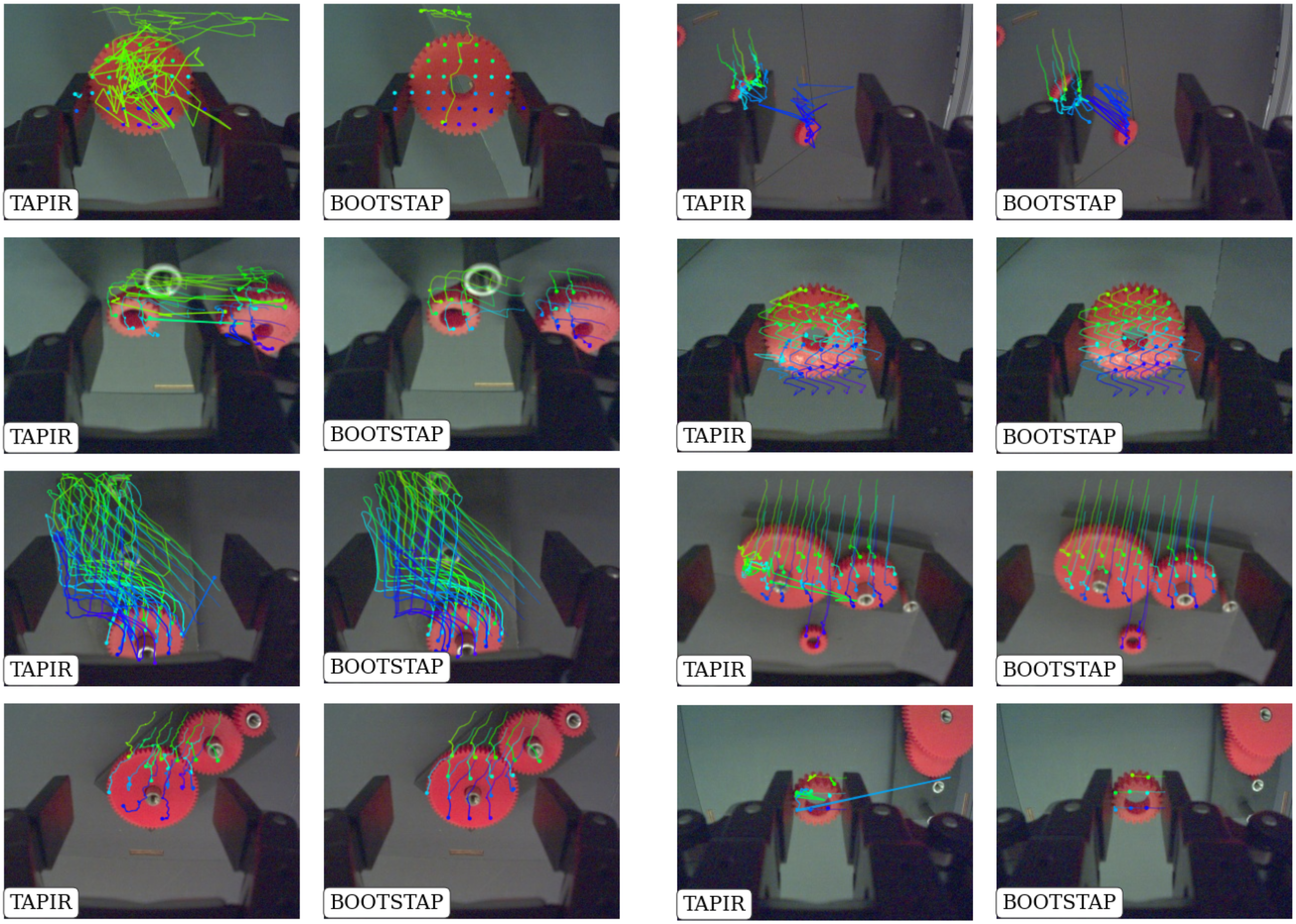}
  \caption{Comparison between TAPIR and BootsTAPIR on the real RoboCAT-NIST dataset. Due to the lack of ground truth, we show the TAPIR prediction and BootsTAPIR prediction in Rainbow tail style side-by-side. On NIST, BootsTAPIR works more consistently on location prediction. Particularly points that were originally predicted as occluded now can be visible.}
  \label{fig:nist}
\end{figure}

Figure~\ref{fig:nist} further illustrates improvements on the RoboCAT-NIST.
Due to the lack of ground truth, sample a grid of points on the red pixels for RoboCAT-NIST.  We display a few examples comparing the predicted tracks between the two models. 
As these are rigid objects, we expect the points to move consistently within each gear; deviations from this are errors.
Due to the lack of texture on the gears and the nontrivial domain gap, the original TAPIR trained on Kubric works poorly here, with many jittery tracks and severe tracking failures.  This is particularly bad for points that are close to occlusion or out of image boundary. 
The bootstrapped model fixes many of these failures: the tracks are much smoother and occlusion predictions become much more accurate.  Results are comparable on Libero, although the motions there are more complicated and unsuitable for a static figure; see our project webpage for video visualizations.

\subsection{Ablations}

We focus on four main areas of ablation: \textbf{data transformations}, \textbf{pseudolabel filtering approaches}, \textbf{training setup}, and \textbf{training data}. To arrive at our final model, we performed ablations on a smaller-scale \base{} setting with our best guesses at the optimal hyperparameter settings. This setting includes two components that we found could be removed without harming performance: It uses an additional mask on the occlusion loss, inspired by FixMatch~\cite{sohn2020fixmatch}, where any occlusion estimate that the teacher is uncertain about ($\max(\sigma(\hat{o}_{\mathcal{T}}[t]), 1- \sigma(\hat{o}_{\mathcal{T}}[t]))<0.6$ is ignored in the loss.
It uses a 3D-ConvNet backbone, which we find provides a slight improvement on DAVIS while harming performance on Kinetics (see Appendix~\ref{app:convnet}), so we remove it for future compatibility with causal TAPIR models.
Finally, \base{} also halves the batch sizes (and proportionally halves the learning rate), and also halves the number of training steps.
We report Average Jaccard on DAVIS using the \emph{strided} mode and on Kinetics using the \emph{q\_first} mode.

\paragraph{Data transformations.} We first investigate the effect of the transformations we apply on inputs and outputs in this setting. 
We respectively ablate: the use of random JPEG augmentations to enforce invariance to various factors of variation (denoted by~\minusaug);
the use of framewise affine transformations on inputs and outputs to enforce equivariance with spatial transformations (denoted by~\minusaffine). 
We also investigate sampling the student queries: recall that in our typical setup, we sample the student query from a distribution which places probability 0.5 on the original teacher query point, and 0.5 on a uniform distribution across visible points.  In~\minusqueries, we always use the teacher's query for the student, and in~\baseuniform, we sample from a purely uniform distribution.
We report the results for each ablation in Table~\ref{tab:consistency_regularization} (a).
We observe that removing JPEG somewhat harms metrics, especially on Kinetics.
In contrast, when ablating affine transformations, we find that performance drops massively across metrics, suggesting overfitting.
Finally, we find that using different query points improves performance compared with~\minusqueries, leading to more accurate position predictions in particular ($\mathbf{< \delta^{x}_{avg}}$ increases from 77.5 to 77.9 on DAVIS \emph{strided} and from 66.8 to 67.7 on Kinetics \emph{q\_first}).  Note, however, on DAVIS in particular, this improvement depends on sampling the original teacher query point more often than the others.

\begin{table}[tb]
\caption{Ablations of model hyperparameters, including (a) ablations of the data transformations and query point strategies, (b) comparisons of the pseudolabel filtering approaches, (c) ablations of the training setup, including the stop gradient, and (d) ablations of the dataset. 
We report Average Jaccard (AJ) across all experiments.
}
\setlength{\tabcolsep}{5pt}
\begin{minipage}[t]{.47\linewidth}
  \subcaptionbox{Data transformations.
  }{
  \label{tab:consistency_regularization}
  \centering
\resizebox{.95\textwidth}{!}{
  \begin{tabular}{lcc}
    \toprule
       & DAVIS & Kinetics \\
Method & \emph{strided} & \emph{q\_first}\\
\midrule
\base           & 65.8  & 54.4  \\
\midrule
\minusaug       & 65.7  & 53.5  \\
\minusaffine    & 54.4  & 44.7  \\
\midrule
\minusqueries   & 65.6  & 53.2  \\
\baseuniform  & 65.6  & 54.3  \\
\bottomrule
  \end{tabular}
}}

\subcaptionbox{Pseudolabels filtering. \base~filters occlusion loss terms based on teacher confidence.}{
  \label{tab:pseudolabels_filtering}
  \centering
\resizebox{.95\textwidth}{!}{
  \begin{tabular}{lcc}
    \toprule
& DAVIS & Kinetics \\
Method & \emph{strided} & \emph{q\_first}\\
\midrule
\base                       & 65.8      & 54.4  \\
\nofiltering                & 65.9      & 54.1  \\
\basecycle                  & 66.1      & 54.3  \\
\bottomrule
  \end{tabular}
  
}}
\vspace{-1em}
\end{minipage}
\hfill
\begin{minipage}[t]{.47\linewidth}
  \subcaptionbox{Training setup.}{
  \label{tab:training_setup}
  \centering
\resizebox{.95\textwidth}{!}{
  \begin{tabular}{lccc}
    \toprule
        & DAVIS & Kinetics \\
Method  & \emph{strided} & \emph{q\_first}\\
\midrule
Full model  & 66.2     & 54.6  \\
\noselfsup  & 65.0     & 52.7  \\
\midrule
\base       & 65.8     & 54.4  \\
\siamese    & 49.8     & 29.6  \\
\bottomrule
  \end{tabular}
  }}
\vspace{0.03cm}

  \subcaptionbox{Training Data.}{
  \label{tab:data_ablation}
  \centering
\resizebox{.95\textwidth}{!}{
  \begin{tabular}{lccc}
    \toprule
        & DAVIS & Kinetics \\
Method  & \emph{strided} & \emph{q\_first}\\
\midrule
\base               & 65.8  & 54.4  \\
\midrule
2-frame clips       & 64.3  & 50.5  \\
6-frame clips       & 63.7  & 50.9  \\
\midrule
1\% of real data    & 66.2  & 54.0 \\
\bottomrule
  \end{tabular}
  
}}
\vspace{-1em}
\end{minipage}
\end{table}

\paragraph{Pseudolabel filtering.}
We next consider the effectiveness of filtering possibly-incorrect teacher tracks and points, with results in Table~\ref{tab:pseudolabels_filtering} (b).  \nofiltering{} removes the filtering that \base{} uses on the occlusion confidence score, which makes little difference in performance on DAVIS, but degrades performance on Kinetics.  \basecycle{} uses cycle-consistency criterion from our full model instead and performs slightly better on DAVIS.  These results suggest that correctly removing bad teacher tracks remains an open problem.

\paragraph{Training setup.}  Table~\ref{tab:training_setup} (c) shows ablations of the overall training setup.  In particular, training for longer with a higher capacity model can improve results, so \noselfsup{} uses an identical training setup to our full model, but removes self-supervised training, instead simply training on Kubric for longer.  We see competitive performance, although self-supervised training still improves by 1.2\% on DAVIS and almost 2\% on Kinetics. \siamese{} shows the effect of removing the EMA and stop-gradient and instead backpropping to both student and teacher models (in this case, using the \base{} setting): we note that performance on real-world datasets collapses as the model finds trivial shortcuts.

\paragraph{Training data.}  We ablate two questions regarding the dataset.  Prior work has argued that simple semi-supervised learning for optical flow performs poorly~\cite{lai2017semi,liu2021learning}; we hypothesize that more temporal context may be the key ingredient to change this story.  To validate this, we re-ran our algorithm using 2- and 6-frame clips from our full dataset.  In Table~\ref{tab:data_ablation} (d), we see that this indeed performs poorly, possibly because the extra frames allow the teacher model to correct more errors.  Interestingly, we also tried training on a 1\% subset of the data, and found that this harms performance on Kinetics but actually improves it on DAVIS.  It's possible that the algorithm begins overfitting to the data, but this may be useful for clean data like DAVIS.  Regardless, it suggests this algorithm can be effective even in situations where less data is available.

\section{Higher Resolution and Public Release}
The publicly released version of BootsTAPIR contains a fix for a minor bug that was present in our previous versions of TAPIR.  Specifically, recall that the data augmentations used for the Kubric dataset include a random axis-aligned crop.  The image cropping mechanism was not pixel-aligned with the transforms used for points, leading to an almost imperceptible error in the track locations. Fixing this bug leads to an improvement in performance for the original TAPIR model, but surprisingly has relatively little effect on BootsTAP performance.  However, we find that the reason isn't because BootsTAP compensates for the bug, but rather, because the bug creates a bias toward tracking foreground objects (the tracks tend to be slightly expanded relative to the underlying objects).  We find we can replicate this bias by altering query points that are very near occlusion edges (1 pixel away) to track the foreground object rather than the background, which we call the ``snap to occluder'' technique.  See Appendix~\ref{app:snap} for details.  

To further tune performance, we also trained on higher-resolution clips, and also longer clips, as we find these improve generalization for real-world applications with longer or higher-resolution videos.  To implement this, we add more `tasks' with different data shapes, using the same multi-task framework (i.e., separate optimizers) as described above.  Specifically, one extra task uses $512\times 512$ Kubric clips (24 frames), trained using the same losses.  We use the hierarchical refinement approach described in the original TAPIR paper, wherein the initialization and one refinement pass is performed at $256\times 256$, and then a further refinement pass is performed at $512\times 512$.  We also use an analogous high-resolution self-supervised task, which also uses 24-frame, $512\times 512$ videos from the same real-world dataset.  Finally, we add 150-frame, $256\times 256$ videos, this time at 30 frames per second.  

\begin{table}[tb]
  \setlength{\tabcolsep}{3pt}
  \caption{Comparison of performance on the TAP-Vid datasets for the released version of BootsTAPIR. Fix refers to the bugfix to coordinates.  Snap refers to the snap-to-occluder bias in the training data.  Data refers to extra training data which has longer clips and higher resolution.
  }
  \label{tab:tapir_perf_release}
  \centering
\resizebox{.99\textwidth}{!}{
  \begin{tabular}{lccc|ccc|ccc}
    \toprule
 & \multicolumn{3}{c|}{Kinetics} & \multicolumn{3}{c|}{DAVIS} & \multicolumn{3}{c}{RGB-Stacking} \\
Method &  AJ & $<\delta^{x}_{avg}$ & OA &  AJ & $<\delta^{x}_{avg}$ & OA &  AJ & $<\delta^{x}_{avg}$ & OA \\
\midrule
CoTracker~\cite{karaev2023cotracker} & 57.3 & 70.6 & 87.5 &  64.8 & 79.1 & 88.7 & 65.9 & 80.6 & 85.0  \\ %
Tuned TAPIR+fix & 60.4 & 72.5 & 88.6 & 64.7 & 76.6 & 89.6 & 70.0 & 82.1 & 89.9 \\
Tuned TAPIR+fix+snap & 59.6 & 71.9 & 88.8 & 63.8 & 75.9 & 89.9 & 67.1 & 80.3 & 87.8 \\
\midrule
BootsTAPIR & 61.4 & 74.2 & 89.7 & 66.2 & 78.1 & 91.0 & 72.4 & 83.1 & 91.2 \\
BootsTAPIR+fix & 61.4 & 74.5 & 89.4 & 66.5 & 78.5 & 90.8 & 75.9 & 85.7 & 93.8 \\
BootsTAPIR+fix+snap & 61.3 & 74.7 & 89.1 & 67.1 & 78.9 & 91.2 & 75.7 & 85.7 & 93.4 \\
BootsTAPIR+fix+snap+data & 62.5 & 74.8 & 89.5 & 67.4 & 79.0 & 91.3 & 77.4 & 86.7 & 93.2 \\
  \end{tabular}
}
\end{table}

\begin{table}[tb]
  \setlength{\tabcolsep}{5pt}
  
  \caption{Comparison of performance under query-first metrics for Kinetics, TAP-Vid DAVIS, and RoboTAP (standard for this dataset).
  }
  \label{tab:tapir_perf_first_release}
  \centering
  \resizebox{.99\textwidth}{!}{
  \begin{tabular}{lccc|ccc|ccc}
    \toprule
 & \multicolumn{3}{c|}{Kinetics} & \multicolumn{3}{c|}{DAVIS} & \multicolumn{3}{c}{RoboTAP} \\ %
Method & AJ & $<\delta^{x}_{avg}$ & OA & AJ & $<\delta^{x}_{avg}$ & OA & AJ & $<\delta^{x}_{avg}$ & OA \\
\midrule
CoTracker~\cite{karaev2023cotracker} & 48.7 & 64.3 & 86.5 & 60.6 & 75.4 & 89.3 & 54.0 & 65.5 & 78.8 \\
Tuned TAPIR+fix & 53.3 & 66.0 & 85.1 & 58.9 & 71.6 & 86.4 & 67.3 & 78.4 & 90.0 \\
Tuned TAPIR+fix+snap & 52.5 & 65.3 & 85.5 & 58.3 & 71.1 & 87.7 & 66.4 & 77.3 & 90.5 \\
\midrule
BootsTAPIR & 54.6 & 68.4 & 86.5 & 61.4 & 73.6 & 88.7 & 64.9 & 80.1 & 86.3 \\
BootsTAPIR+fix & 54.7 & 68.5 & 86.3 & 61.6 & 74.1 & 89.0 & 65.7 & 80.5 & 87.2 \\
BootsTAPIR+fix+snap & 54.5 & 68.8 & 86.3 & 61.8 & 74.3 & 89.1 & 63.5 & 81.1 & 84.2 \\
BootsTAPIR+fix+snap+data & 55.8 & 68.8 & 86.6 & 62.4 & 74.6 & 89.6 & 69.2 & 81.3 & 89.5 \\
\bottomrule
  \end{tabular}
  }
\end{table}

Tables~\ref{tab:tapir_perf_release} and~\ref{tab:tapir_perf_first_release} show our results.  Note that CoTracker implemented its own data augmentation algorithms and is not affected by the same bug.  We see that ``snap to occluder'' harms TAPIR performance, but improves BootsTAP performance.  One possible interpretation is that the snapping is compensating for a particular bias in the bootstrapping toward tracking background.  This may be because background is easier to track, especially relative to thin objects.  In a bootstrapping framework, the model’s reliable predictions that follow background become self-reinforcing, whereas unreliable predictions for thin foreground objects, are not.  Therefore, they tend to get lost over time.  Finding more principled solutions to this issue is an interesting area for future work.

The extra training data, however, leads to a non-trivial boost in performance.  Note that, for tables~\ref{tab:tapir_perf_release} and~\ref{tab:tapir_perf_first_release}, all evaluation videos are still at $256\times 256$, and unlike many prior methods we do not upsample them before creating the feature representation.  To assess the impact of increased \textit{evaluation} resolution, we also performed evaluation on $512\times 512$ videos.  Table~\ref{tab:tapir_perf_hires} shows results.  We see that performance improves by 1.2\% on Kinetics and 2.8\% on DAVIS, the best reported performance on this dataset by a wide margin.  Surprisingly, we found that further resolution at test-time did not improve results, suggesting another interesting area for future work.

\begin{table}[tb]
  \setlength{\tabcolsep}{5pt}
  
  \caption{Comparison of performance for high-resolution setting.
  }
  \label{tab:tapir_perf_hires}
  \centering
  \begin{tabular}{lccc|ccc}
    \toprule
 & \multicolumn{3}{c|}{Kinetics} & \multicolumn{3}{c}{DAVIS}  \\ %
Method & AJ & $<\delta^{x}_{avg}$ & OA & AJ & $<\delta^{x}_{avg}$ & OA \\
\midrule
BootsTAPIR& 62.5 & 74.8 & 89.5 & 67.4 & 79.0 & 91.3 \\
BootsTAPIR+hires& 63.7 & 76.0 & 88.4 & 70.2 & 81.2 & 91.2 \\
\bottomrule
  \end{tabular}
\end{table}

RoboTAP~\cite{vecerik2023robotap} pointed out that point tracking can be very useful in an online setting, e.g. when used as a signal to control agents in real time.  It remains straightforward to extend BootsTAPIR to the online setting: the only temporal dependency of the model is in the 1D convolutions in the iterative refinements, so these can be directly converted into causal convolutions to create a causal model.  We trained this model using the full training setup for the release model, including the extra high-resolution, long-clip data.  Table~\ref{tab:tapir_perf_causal} shows results.  We see an overall 4.6\% improvement on Kinetics and a 3.0\$ improvement on DAVIS, in both cases using the query-first evaluation procedure.  

\begin{table}[tb]
  \setlength{\tabcolsep}{5pt}
  
  \caption{Causal model performance.
  }
  \label{tab:tapir_perf_causal}
  \centering
  \begin{tabular}{lccc|ccc}
    \toprule
 & \multicolumn{3}{c|}{Kinetics \emph{q\_first}} & \multicolumn{3}{c}{DAVIS \emph{q\_first}}  \\ %
Method & AJ & $<\delta^{x}_{avg}$ & OA & AJ & $<\delta^{x}_{avg}$ & OA \\
\midrule
Causal TAPIR & 51.5 & 64.4 & 85.2 & 56.7 & 70.2 & 85.7 \\
Causal BootsTAPIR & 55.1 & 67.5 & 86.3 & 59.7 & 72.3 & 86.9 \\
\bottomrule
  \end{tabular}
\end{table}

We also perform experiments on the point tracks in the Perception Test~\cite{patraucean2024perception} validation set, a challenging dataset of point tracks annotated on videos of unusual situations filmed by participants.  Results are shown in Table~\ref{tab:tapir_perception_test}; we see a similar magnitude gap over prior results.

\begin{table}[tb]
  \setlength{\tabcolsep}{5pt}
  
  \caption{Performance on Perception Test relative to TAPIR.
  }
  \label{tab:tapir_perception_test}
  \centering
  \begin{tabular}{lccc}
    \toprule
Method & Overall & Static & Dynamic \\
\midrule
TAPIR &  55.7 & 57.4 & 46.3 \\
BootsTAPIR & 59.6 & 61.3 & 49.7 \\
\bottomrule
  \end{tabular}
\end{table}

As a final note, we performed informal benchmarking of our model using an A100 and the latest JAX compiler.  We found that after compilation, BootsTAPIR can perform inference of $10,000$ points on a $256\times 256$, 50-frame video in 5.6 seconds.  Furthermore, the causal model can track 400 points on a $256\times 256$ video at 30.1 frames per second.

\section{Conclusion}
In this work we presented an effective method for leveraging large scale, unlabeled data for improving TAP performance. We have demonstrated that a straightforward application of consistency principles, namely invariance to query points and non-spatial corruptions, and equivariance to affine transformations, enable the model to continue to improve on unlabeled data. Our formulation avoids more complex priors such as spatial smoothness of motion or temporal smoothness of tracks that are used in many prior works. In fact, our formulation bears similarities to baselines for two-frame, self-supervised optical flow that are considered too ``unstable'' to be effective (c.f. Fig. 2(a) in ``Flow Supervisor''~\cite{im2022semi}). Yet in our multi-frame approach, we ultimately surpass the state-of-the-art performance by a large margin. We find little evidence of model `overfitting' to its own biases in ways that cause performance to degrade with long training like in other work \cite{sun2024refining}. Instead, we find that performance continues to improve for as long as we train the model.  Our work does have some limitations: training remains computationally expensive. Furthermore, our estimated correspondence is a single point estimate throughout the entire video, which means we cannot elegantly handle duplicated or rotationally-symmetric objects where the actual correspondence is ambiguous.  Nevertheless, our approach demonstrates that it is possible to better bridge the sim-to-real gap using self-supervised learning.

\bibliographystyle{splncs04}
\bibliography{shortstrings, main}

\newpage
\appendix

\section{Summary of the approach}
\label{app:recap}

We summarize notation and computation of our self-supervised loss in Algorithm~\ref{alg:cap}.

\input{method_pseudocode}

\section{Implementation details}

\subsection{Distribution over transformations}
\label{appendix:affine}

We design transformations of the inputs to enforce equivariance of the predictions with {\em realistic} spatial transformations. At a high level, we intend for the transformations to mimic the effects of additional, simple and plausible camera motion and zooming on the video. Hence, our transformations should vary smoothly in time; they should cover a reasonable ratio of the original video content; and aspect ratio should be roughly preserved.

We define a family of frame-wise affine transformations that has these properties, and a procedure to sample these randomly.
Essentially, we sample top-left crop coordinates and crop dimensions for each frame in the video; where coordinates and dimensions are computed as interpolations between values sampled for the start and end frames from a distribution that achieves the desired coverage and aspect ratios.

More formally, we first sample a pair of spatial dimensions $(H_0, W_0)$ for the start frame as follows. We sample an area $A$ uniformly over $[0.6, 1.0]$. Next we sample values $a^1, a^2 \sim \mathcal{U}([A, 1])$ and derive random height value by averaging them $h = \frac{a^1 + a^2}{2}$ and width value $w = \frac{A}{h}$; and finally, we multiply these values by the input's original shape $(H, W)$.
This gives us a pair of spatial dimensions biased towards aspect ratios close to 1, and covering an area between $60\%$ and $100\%$ of the original input.
We proceed the same way to sample a pair of spatial dimensions$(H_{T-1}, W_{T-1})$ for the end frame.

Next, we uniformly sample a pair of top-left corner coordinates $(C_0^x, C_0^y)$ for the start frame, such that a crop of dimensions $(H_0, W_0)$ can be extracted within the frame.
We proceed the same way to sample a pair of spatial coordinates $(C_{T-1}^x, C_{T-1}^y)$, given $(H_{T-1}, W_{T-1})$, for the end frame.

We then interpolate linearly on one hand between the start and end spatial dimensions; and on the other hand between the start and end top-left corner coordinates. Let $t \in \{0, ..., T-1\}$ be a frame index. Calling $\alpha_t = \frac{t}{T-1}$, we define:

\begin{align}
    h_t = (1 - \alpha_t) * H_0 + \alpha_t * H_{T-1} \\
    w_t = (1 - \alpha_t) * W_0 + \alpha_t * W_{T-1} \\
    c_t^x = (1 - \alpha_t) * C_0^x + \alpha_t * C_{T-1}^x \\
    c_t^y = (1 - \alpha_t) * C_0^y + \alpha_t * C_{T-1}^y.
\end{align}

\noindent This gives us parameters of scaling parameters $(h_t, w_t)$ and translation parameters $(c_t^x, c_t^y)$ vary linearly over time.
Finally, our frame-wise affine transformations $\Phi = \{\Phi_t\}_t$ are defined as follows:

\begin{equation}
    \forall t, \Phi_t: (x, y) \mapsto \Big(\frac{w_t}{W} * x + c_t^x, \frac{h_t}{H} * y + c_t^y\Big),
\end{equation}

\noindent We refer to the distribution resulting from our sampling procedure as $\mathcal{D}_\Phi$. Given a query point coordinates $Q = (q, t)$ and input frames $\{X_t\}_t$, the corresponding transformation is applied with:

\begin{equation}
    Q' = (\Phi_t(q), t); \;\;\;  \forall t, X'_t = \text{resample}(X_t, \Phi_t),
\end{equation}

\noindent where $\text{resample}( \cdot, \Phi_t)$ consists in scaling its input frame to resolution $(h_t, w_t)$ using bilinear interpolation and placing it within a zero-valued array of shape $(H,W)$ such that its top-left corner in the array is at coordinates $(c_t^x, c_t^y)$. We note that in our approach, this transformation is performed after augmenting each frame, \ie on $a(X_t)$.

\subsection{Training details}
\label{appendix:details}
We train for 200,000 iterations on 256 nVidia A100 GPUs, with a batch size of 4 Kubric videos and 2 real videos per device.  The extra layers consist of 5 residual blocks on top of the backbone (which has stride 8, 256 channels), each of which consists of 2 sequential $3 \times 3$ convolutions with a channel expansion factor of 4, which is then added to the input.  We use a cosine learning rate schedule with 1000 warmup steps and a peak learning rate of 2e-4.  We found it improved stability to reduce the learning rate for the PIPs mixer steps relative to the backbone by a factor of 5. We keep all other hyperparameters the same as TAPIR. 

\subsection{Libero finetuning}
\label{app:libero}
We compare results on Libero, using the gripper view, which contains large and difficult motions.  Qualitative results show that BootsTAP trained on internet videos as described improves results substantially.  However, since there's a large domain gap between Libero data and internet videos, it's natural to ask whether performance can be improved by further self-supervised training on the Libero dataset.

We use the full set of demonstrated trajectories in the dataset for all tasks, again using only the gripper view.  We begin with the model trained as described in the main paper, and then further train it for another 50K steps using three tasks jointly: Kubric, internet videos, and Libero videos, again using separate optimizers for each and summing updates across tasks.  We use an update weight of 0.2 for both self-supervised tasks, and keep all other parameters the same between Libero and the internet video tasks.  We see that this approach further improves results despite having no labels: the model can track with surprisingly high fidelity over large changes in scale and viewpoint.  See the attached html file for visualizations.

\subsection{Snap-to-occluder}
\label{app:snap}
We aim to slightly modify the training objective to bias TAPIR to track foreground objects rather than background, to counteract the tendency of bootstrapped models to track background.  The Kubric data loader works by sampling query pixels randomly (biased toward objects), and then computing the full track by back-projecting into the relevant object's local coordinate system.  We first modify the procedure by preventing the model from sampling pixels on the 'back side' of an occlusion boundary: this is defined as any pixel with a neighboring pixel (within a 3x3 square) which is less than 95\% of the pixel's depth.  After tracking points, we identify query points that are on the 'front side' of an occlusion boundary: that is, any neighboring pixel which is more than 105\% of the depth of the query point.  If such pixels exist, with 50\% probability we randomly choose one such pixel and replace the query point with it.  Therefore, in a small fraction of cases, the model will receive a query point on the background but need to track the foreground object instead.

\section{Comparison with and without a 3D ConvNet Backbone}
\label{app:convnet}

Recall that TAPIR extracts features using a ResNet, with a final feature map of dimension 256 at stride 8 (although it uses an earlier feature map as well at stride 4).  The architecture is similar to a ResNet-18, and therefore has relatively little capacity to learn about the full diversity of objects in the world.  Therefore, we add extra capacity: 5 more ResNet layers consisting of a LayerNorm, a $3\times 3$ convolution, followed by a GeLU, followed by another $3 \times 3$ convolution which is added to the input of the layer.  Like with TAPIR, our full model applies the feature extractor independently on every frame, meaning that the model cannot use temporal cues for feature extraction.  Is this choice optimal?  Intuitively, we might expect motion to provide segmentation cues that could enable better matching.  Therefore, we develop an alternative model which adds a simple 3D ConvNet into the backbone: specifically, we convert the first convolution of each residual block layer into a $3\times 3\times 3$, giving the features a temporal receptive field of 21 frames.

We report results Table~\ref{tab:tsm}.  
We observe that this yields a slight performance increase on TAP-Vid DAVIS (\emph{strided} evaluation), and in particular, slightly improves the position accuracy, although it harms occlusion accuracy.  However, it significantly degrades performance on Kinetics (\emph{query\_first}).  It's possible that the model struggles more with the cuts or camera shake present in Kinetics.  Hence, we keep a 2D backbone for the final model, although the optimal model may depend on the desired downstream application.

\begin{table}[tb]
  \setlength{\tabcolsep}{5pt}
  \caption{Architectures. We compare a 2D backbone and a 3D backbone using Temporal Shift Modules (TSM) to aggregate information locally over time.}
  \label{tab:tsm}
  \centering
  \begin{tabular}{lccc|ccc}
    \toprule
 & \multicolumn{3}{c}{TAP-Vid DAVIS - \emph{strided}} & \multicolumn{3}{c}{Kinetics - \emph{query\_first}} \\ %
 & AJ & $<\delta^{x}_{avg}$ & OA & AJ & $<\delta^{x}_{avg}$ & OA \\
\midrule
2D ConvNet Backbone & 66.2 & 78.1 & 91.0 & 54.6 & 68.4 & 86.5  \\
3D ConvNet Backbone & 66.3 & 78.4 & 90.7 & 54.0 & 68.0 & 85.0 \\
\bottomrule
  \end{tabular}
\end{table}

\section{Evaluation Datasets}
\label{app:evals}

\noindent \textbf{TAP-Vid-Kinetics} contains videos collected from the Kinetics-700-2020 validation set \cite{carreira2017quo} with original focus on video action recognition. This benchmark contains 1K internet videos of diverse action categories, approximately 10 seconds long, including many challenging elements such as shot boundaries, multiple moving objects, dynamic camera motion, cluttered background and dark lighting conditions.  Each video contains $\sim$26 tracked points on average, obtained from careful human annotation.  

\noindent \textbf{TAP-Vid-DAVIS} contains 30 real-world videos from DAVIS 2017 validation set~\cite{perazzi2016benchmark}, a standard benchmark for video object segmentation, which was extended to TAP. 
Each video contains $\sim$22 point tracks using the same human annotation process as TAP-Vid-Kinetics. 

\noindent \textbf{TAP-Vid-RGB-Stacking} contains 50 synthetic videos generated with Kubric~\cite{greff2022kubric} which simulate a robotic stacking environment. Each video contains 30 annotated point tracks and has a duration of 250 frames.

\noindent \textbf{RoboTAP} contains 265 real world Robotics Manipulation videos with on average $\sim$272 frames and $\sim$44 annotated point tracks per video \cite{vecerik2023robotap}. 
These videos are even longer, with textureless and symmetric objects that are far out-of-domain for both Kubric and the online lifestyle videos that we use for self-supervised learning. %

\noindent \textbf{RoboCAT-NIST} is a subset of the data collected for RoboCat~\cite{bousmalis2023robocat}. Inspired by the NIST benchmark for robotic manipulation \cite{kimble2020benchmarking}, it includes gears of varying sizes (small, medium, large) and a 3-peg base, introduced for a systematic study of insertion affordance. All videos are collected by human teleoperation. It includes robot arms operating and inserting gears, which are a particularly challenging case due to the rotational symmetry and lack of texture. In this work, we processed videos to 64 frames long with 222 $\times$ 296 resolution. This dataset is mainly for demonstration purpose, there are no human groundtruth point tracks.

\noindent \textbf{Libero}~\cite{liu2024libero} is a dataset where point tracking has already proven useful for robotic manipulation~\cite{abbeel}.  It includes demos of a human-driven robot arm performing a wide variety of tasks in a synthetic environment, intended for use in imitation learning.  Sequences are variable length at $128 \times 128$ resolution and has no ground truth tracks.

\subsection{Evaluation metrics}

We use three evaluation metrics same as proposed in \cite{doersch2022tap}. (1) $\mathbf{< \delta^{x}_{avg}}$ is the average position accuracy across 5 thresholds for $\delta$: 1, 2, 4, 8, 16 pixels. For a given threshold $\delta$, it computes the proportion of visible points (not occluded) that are closer to the ground truth than the respective threshold. (2) \textbf{Occlusion Accuracy (OA)} is the average binary classification accuracy for the point occlusion prediction at each frame. (3) \textbf{Average Jaccard (AJ)} combines the two above metrics and is typically considered the target for this benchmark.  It is the average Jaccard score across the same thresholds as $< \delta^{x}_{avg}$. Jaccard at $\delta$ measures both occlusion and position accuracy. It is the fraction of ‘true positives’, i.e., points within the threshold of any visible ground truth points, divided by ‘true positives’ plus ‘false positives’ (points that are predicted visible, but the ground truth is either occluded or farther than the threshold) plus ‘false negatives’ (groundtruth visible points that are predicted as occluded, or where the prediction is farther than the threshold). 

For TAP-Vid datasets, evaluation is split into \textit{strided mode} and \textit{query-first mode}. Strided mode samples query points every 5 frames on the groundtruth tracks when they are visible. Query points can be any time in the video hence it tests the model prediction power both forward and backward in time. Query-first mode samples query points only when they are first time visible and the evaluation only measures tracking accuracy in future frames.

\end{document}

%% file: method_pseudocode.tex
\begin{algorithm}[H]

\caption{BootsTAP self-supervised loss. Notation: \\
$\mathcal{U}(D)$ refers to the uniform distribution over domain $D$;  \\
we denote queries as $Q = (q, t)$ where $q$ is x/y coordinates and $t$ is a frame index.  In a slight abuse of notation, we call $\Phi_t$ the transformation and the mapping that transforms coordinates and leaves other model outputs unchanged.
}\label{alg:cap}
\begin{algorithmic}
\Require  \\
$X$ -- video of shape $T \times H \times W \times C$  \\
$f$ -- model \\
$\Theta$, $\xi$ -- student parameters, teacher parameters \\
$\mathcal{A}, \mathcal{D}_\Phi$ -- distribution over augmentations, distribution over transformations \\
$\mathcal{V} \mapsto \mathcal{D}_{\mathcal{V}}$ -- mapping that maps a set of points $\mathcal{V}$ to a distribution $\mathcal{D}_{\mathcal{V}}$ over $\mathcal{V}$\\
$\delta$, $\delta_{cycle}$ -- threshold values for uncertainty target definition and cycle-consistency filtering criterion\\
$d(\cdot, \cdot)$ -- distance function  \\

\State Uniformly sample teacher query points $Q_1 \sim \mathcal{U}([0, H) \times [0, W) \times \llbracket 0, T-1\rrbracket)$.

\State Sample augmentation $a \sim \mathcal{A}$ and a frame-wise affine transformation $\Phi = \{\Phi_t\}_t \sim \mathcal{D}_\Phi$.

\State Augment and transform each frame to form $X'$: $\forall t, X'_t \gets \text{resampling}(a(X_t), \Phi_t)$.

\State For each query point $Q_1$:
\State \hspace{1em} Predict tracks and occlusions with teacher model: $\left\{ \hat{p}_\mathcal{T}[t], \hat{o}_\mathcal{T}[t] \right\}_t \gets f(X , Q_1; \xi)$.

\State \hspace{1em} Derive pseudo-labels from teacher predictions with:
\begin{equation*}
    p_{\mathcal{T}}[t] = \hat{p}_{\mathcal{T}}[t] \quad ; \quad 
    o_{\mathcal{T}}[t] = \mathbbm{1}(\hat{o}_{\mathcal{T}}[t] > 0); \quad 
    u_{\mathcal{T}}[t] = \mathbbm{1}(d(p_{\mathcal{T}}[t], \hat{p}_{\mathcal{S}}[t]) > \delta)
\end{equation*}

\State \hspace{1em}  Calling $\mathcal{V}$ the set of visible points along the teacher trajectory, 
\State \hspace{1em} sample $Q_2=(q_2, t_2) \sim \mathcal{D}_{\mathcal{V}}$.

\State \hspace{1em} Transform query points: $Q'_2 \gets (\Phi_{t_2}(q_2), t_2)$.

\State \hspace{1em} Predict tracks with the student model and transform predicted coordinates with\\
\hspace{2em} the inverse of $\Phi_t$:
$\left\{\hat{p}_\mathcal{S}[t], \hat{o}_\mathcal{S}[t], \hat{u}_\mathcal{S}[t] \right\}_t \gets \Phi_t^{-1}(f(X', Q'_2; \Theta))$.

\State \hspace{1em} Compute masks used to filter out loss terms (when $t_1$ and $t_2$ differ):
\begin{equation*}
    m_{cycle} =  \mathbbm{1} \left( d(\hat{p}_{\mathcal{S}}[t_1], q_1) < \delta_{cycle} \right) \quad * \quad \mathbbm{1} \left(\hat{o}_{\mathcal{S}}[t_1] \le 0\right)
\end{equation*}

\State \hspace{1em} Compute the loss:
\begin{equation*}
    \mathcal{L}_{SSL} = m_{cycle} * \frac{1}{T} \sum_t  \ell_{ssl}^t
\end{equation*}
\hspace{1em} where $\ell_{ssl}^t$ is the self-supervised TAPIR loss term for $t$.

\end{algorithmic}
\end{algorithm}